%% file: 0_main.tex
\documentclass[10pt,twocolumn,letterpaper]{article}

\usepackage[pagenumbers]{cvpr}

\input{preamble}

\definecolor{cvprblue}{rgb}{0.21,0.49,0.74}
\usepackage[pagebackref,breaklinks,colorlinks,citecolor=cvprblue]{hyperref}
\newcommand{\sysname}{StableVITON}
\usepackage{cuted}
\usepackage{capt-of}
\usepackage{pifont}

\title{StableVITON: Learning Semantic Correspondence \\ with Latent Diffusion Model for Virtual Try-On}

\author{Jeongho Kim\quad\quad Gyojung Gu\quad\quad Minho Park\quad\quad Sunghyun Park\quad\quad Jaegul Choo\\
KAIST, Daejeon, South Korea\\
{\tt\small \{rlawjdghek, gyojung.gu, m.park, psh01087, jchoo\}@kaist.ac.kr}
}

\begin{document}
\maketitle
\begin{strip}
    \vspace{-1.5cm}
    \centering
    \includegraphics[width=\textwidth]{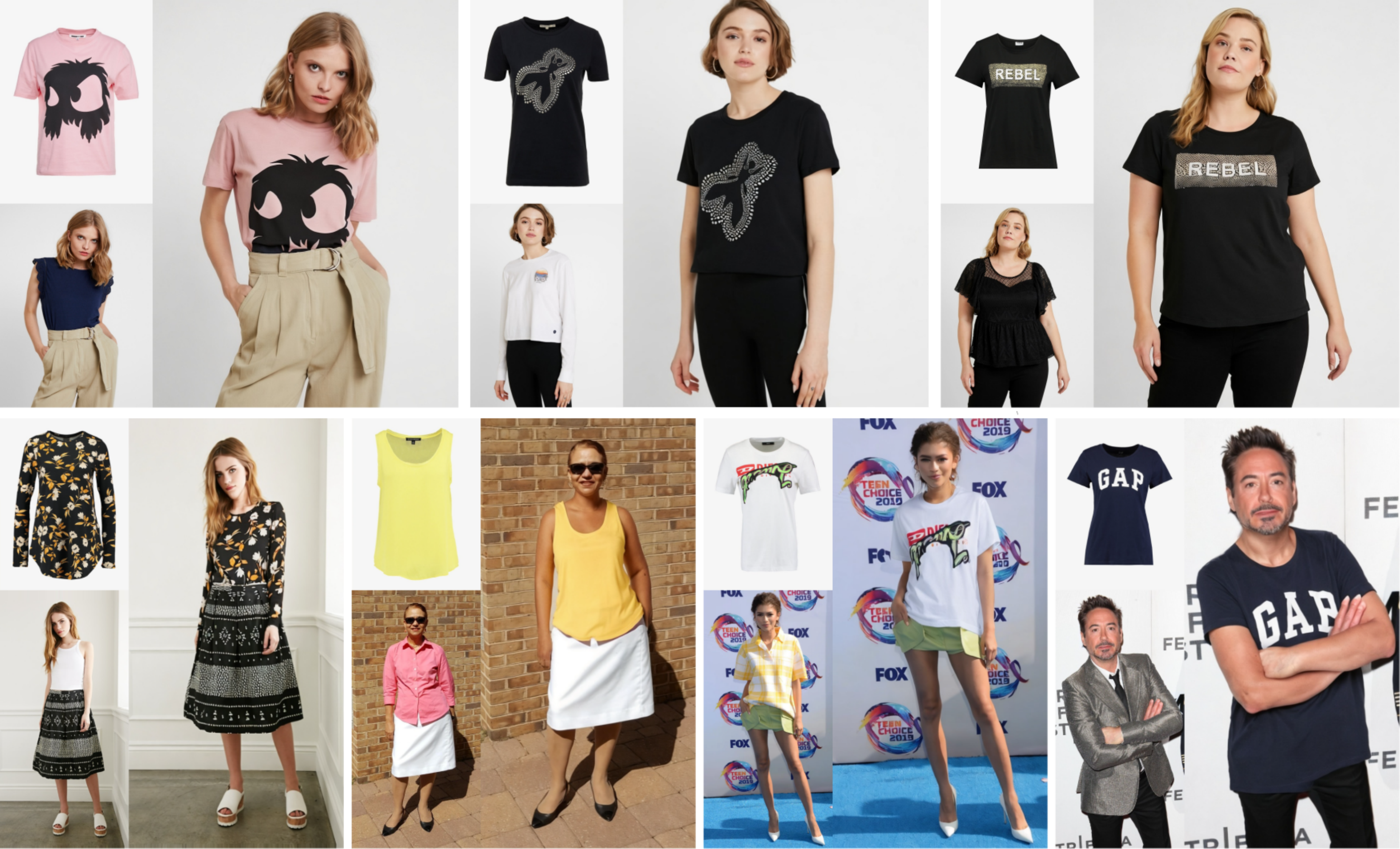}
    \vspace{-0.65cm}
    \captionof{figure}{Generated results of~\sysname: VITON-HD (the first row), SHHQ-1.0 (the first two images in the second row), and web-crawled images (the last two images in the second row). All results are generated using~\sysname~trained on VITON-HD dataset.}
    \vspace{-0.3cm}
\end{strip}
\input{0_abstract}    
\input{1_intro}
\input{2_related}
\input{3_method}
\input{4_experiment}
\input{5_ablation}
\input{6_conclusion}

{
    \small
    \bibliographystyle{ieeenat_fullname}
    \bibliography{main}
}
\input{7_suppl}

\end{document}

%% file: preamble.tex
\usepackage[dvipsnames]{xcolor}

\usepackage{lipsum}

%% file: 0_abstract.tex
\begin{abstract}
\vspace{-0.2cm}
Given a clothing image and a person image, an image-based virtual try-on aims to generate a customized image that appears natural and accurately reflects the characteristics of the clothing image. 
In this work, we aim to expand the applicability of the pre-trained diffusion model so that it can be utilized independently for the virtual try-on task.
The main challenge is to preserve the clothing details while effectively utilizing the robust generative capability of the pre-trained model. 
In order to tackle these issues, we propose~\sysname, learning the semantic correspondence between the clothing and the human body within the latent space of the pre-trained diffusion model in an end-to-end manner. 
Our proposed zero cross-attention blocks not only preserve the clothing details by learning the semantic correspondence but also generate high-fidelity images by utilizing the inherent knowledge of the pre-trained model in the warping process.
Through our proposed novel attention total variation loss and applying augmentation, we achieve the sharp attention map, resulting in a more precise representation of clothing details.
\sysname~outperforms the baselines in qualitative and quantitative evaluation, showing promising quality in arbitrary person images.
Our code is available at \url{https://github.com/rlawjdghek/StableVITON}.
\end{abstract}

%% file: 1_intro.tex
\section{Introduction}
The objective of an image-based virtual try-on is to dress a given clothing image on a target person image.
Most of the previous virtual try-on approaches~\cite{han2018viton,wang2018toward,yang2020towards,ge2021parser,issenhuth2020not,choi2021viton,lee2022high,xie2023gp} leverage paired datasets consisting of clothing images and person images wearing those garments for training purposes.
These methods typically include two modules: (1) a warping network to learn the semantic correspondence between the clothing and the human body, and (2) a generator that fuses the warped clothing and the person image.

Despite achieving significant advancements, previous methods~\cite{choi2021viton,lee2022high,xie2023gp,gou2023taming} still have limitations in achieving generalizability, particularly in maintaining the complex background in an arbitrary person image.
The nature of matching clothing and individuals in the virtual try-on dataset~\cite{han2018viton,choi2021viton,morelli2022dress} makes it challenging to collect data in diverse environments~\cite{neuberger2020image}, which in turn leads to limitations in the generator's generative capability.

Meanwhile, recent advancements in large-scale pre-trained diffusion models~\cite{ramesh2021zero,rombach2022high,saharia2022photorealistic} have led to the emergence of downstream tasks~\cite{ruiz2023dreambooth,li2023gligen,morelli2023ladi,gou2023taming,zhang2023adding,gafni2022make,zhang2023inversion} that control the pre-trained diffusion models for task-specific image generation.
Thanks to the powerful generative ability, several works~\cite{zhang2023adding, lugmayr2022repaint} have succeeded in synthesizing high-fidelity human images using the prior knowledge of the pre-trained models, which signifies the potential for extension to the virtual try-on task.

In this paper, we aim to expand the applicability of the pre-trained diffusion model to provide a standalone model for the virtual try-on task.
In the effort to adapt the pre-trained diffusion model for virtual try-on, a significant challenge is to preserve the clothing details while harnessing the knowledge of the pre-trained diffusion model. 
This can be achieved by learning the semantic correspondence between clothing and the human body using the provided dataset.
Recent research~\cite{morelli2023ladi, gou2023taming} that has employed pre-trained diffusion models in virtual try-on has shown limitations due to the following two issues: (1) insufficient spatial information available for learning the semantic correspondence~\cite{morelli2023ladi}, and (2) the pre-trained diffusion model not being fully utilized, as it pastes the warped clothing in the RGB space, relying on external warping networks as previous approaches~\cite{choi2021viton,lee2022high,xie2023gp,yang2020towards,ge2021parser} for aligning the input condition. 

\begin{figure}[t!]
    \centering
    \includegraphics[width=1\linewidth]{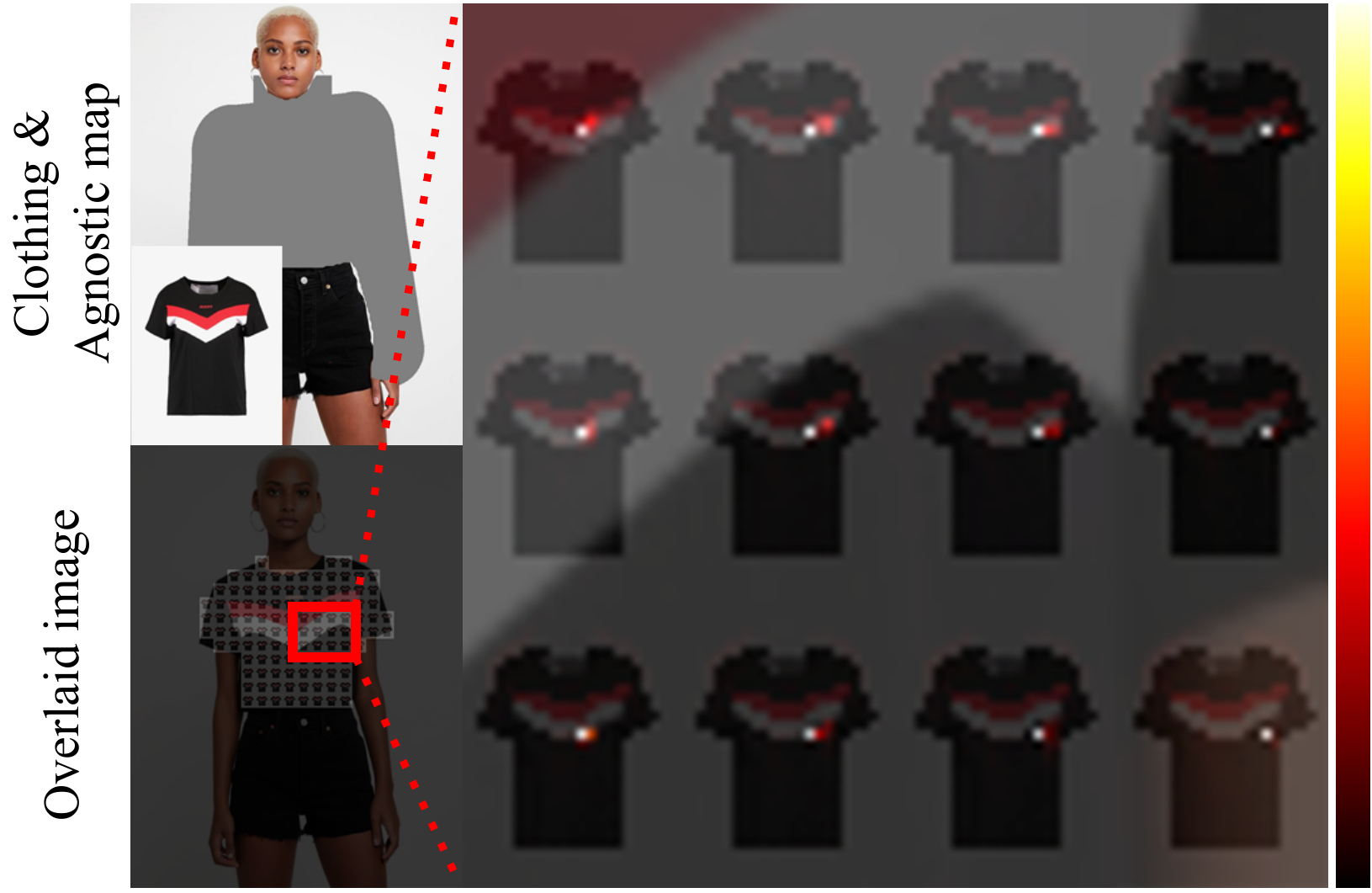}
    \vspace{-0.4cm}
    \caption{Visualization of the semantic correspondence learned by our~\sysname. We overlay the attention map for the clothing regions onto the generated images for visualization.}
    \vspace{-0.3cm}
    \label{fig:correspondence}
\end{figure}

To overcome these issues, we propose~\sysname, which learns the semantic correspondence between the clothing and the human body within the latent space of the pre-trained diffusion model.
To incorporate the spatial information of the clothing for learning semantic correspondence, we introduce an encoder~\cite{zhang2023adding} that takes clothing as input and conditions the U-Net with the intermediate features of the encoder via zero cross-attention blocks. 
Warping through the zero cross-attention block in a pre-trained diffusion model has the following two advantages: (1) preserving the clothing details by learning the semantic correspondence; (2) synthesizing high-fidelity images by leveraging the pre-trained models' inherent knowledge about humans in the warping process.
As shown in Fig.~\ref{fig:correspondence}, the attention mechanism in the latent space performs patch-wise warping by activating each token corresponding to clothing alignment within the generation region.

To further sharpen attention maps, we propose a novel attention total variation loss and apply the augmentation, which yields improved preservation of clothing details.
By not impairing the pre-trained diffusion model, this architecture generates high-quality images even when images with complex backgrounds are provided, only using an existing virtual try-on dataset. Our extensive experiments show that~\sysname~outperforms the existing virtual try-on method by a large margin. 
In summary, our contributions are as follows:
\begin{itemize}
    \item Our proposed~\sysname, to the best of our knowledge, is the first end-to-end virtual try-on method finetuned on the pre-trained diffusion model without an independent warping process.
    \item We propose a zero cross-attention block, which learns semantic correspondence between the clothing and the human body, to condition the intermediate features from the spatial encoder. 
    \item We propose a novel attention total variation loss and apply augmentation for further precise semantic correspondence learning. 
    \item \sysname~shows state-of-the-art performance over existing virtual try-on models in both qualitative and quantitative results. Moreover, through the evaluation of a trained model on multiple datasets,~\sysname~demonstrates its promising quality in a real-world setting.
\end{itemize}

%% file: 2_related.tex
\section{Related Work}
\noindent\textbf{GAN-based Virtual Try-On.}
To properly try-on the given clothing image to the target person, existing approaches~\cite{choi2021viton,lee2022high,xie2023gp,ge2021parser} based on generative adversarial network (GAN) have attempted to address the virtual try-on problem using a two-stage strategy: (1) deforming the clothing to the proposal region and (2) fusing the warped clothing via try-on generator based on GAN. 
In order to achieve precise clothing deformation, previous methods~\cite{lee2022high,ge2021parser,han2019clothflow,bai2022single,xie2023gp} leverage a trainable network that estimates a dense flow map~\cite{zhou2016view} to deform the clothing to the human body.
At the same time, several approaches~\cite{yang2020towards,lee2022high,xie2023gp,choi2021viton,issenhuth2020not,ge2021parser} have been attempted to address the misalignment between the warped clothing and the human body, such as using a normalization~\cite{choi2021viton} or distillation~\cite{issenhuth2020not, ge2021parser}.
However, the existing approaches still are not generalized well, leading to significant performance degradation in arbitrary person images with complex backgrounds.
In this paper, we effectively address such issues by proposing a method that leverages the powerful generation ability of the pre-trained model. 

\noindent\textbf{Diffusion-based Virtual Try-On.}
Due to the remarkable generative capabilities, research on virtual try-on has extensively discussed the application of the diffusion models.
While TryOnDiffusion~\cite{zhu2023tryondiffusion} introduces an architecture for try-on using two U-Nets, this method requires a large-scale and challenging-to-collect dataset, consisting of image pairs of the same person wearing the same clothing in two different poses.
Therefore, much recent research has shifted their focus towards using the prior of a large-scale pre-trained diffusion models~\cite{ho2020denoising,rombach2022high,radford2021learning,yang2023paint} in the virtual try-on task. 
LADI-VTON~\cite{morelli2023ladi} represents the clothing as pseudo-words, and DCI-VTON~\cite{gou2023taming} applies a warping network to input the clothing as conditions for the pre-trained diffusion models.
While both models deal with background-related issues, they suffer from preserving high-frequency details due to the excessive loss of spatial information from the CLIP encoder~\cite{morelli2023ladi} and drawbacks such as incorrectly warped clothing inherited from the independent warping network~\cite{gou2023taming}.
On the other hand, we propose to condition the intermediate feature maps of a spatial encoder through zero cross-attention block, which allows for using the prior knowledge of the pre-trained model in the warping process.

%% file: 3_method.tex
\begin{figure*}[t!]
    \centering
    \includegraphics[width=1\linewidth]{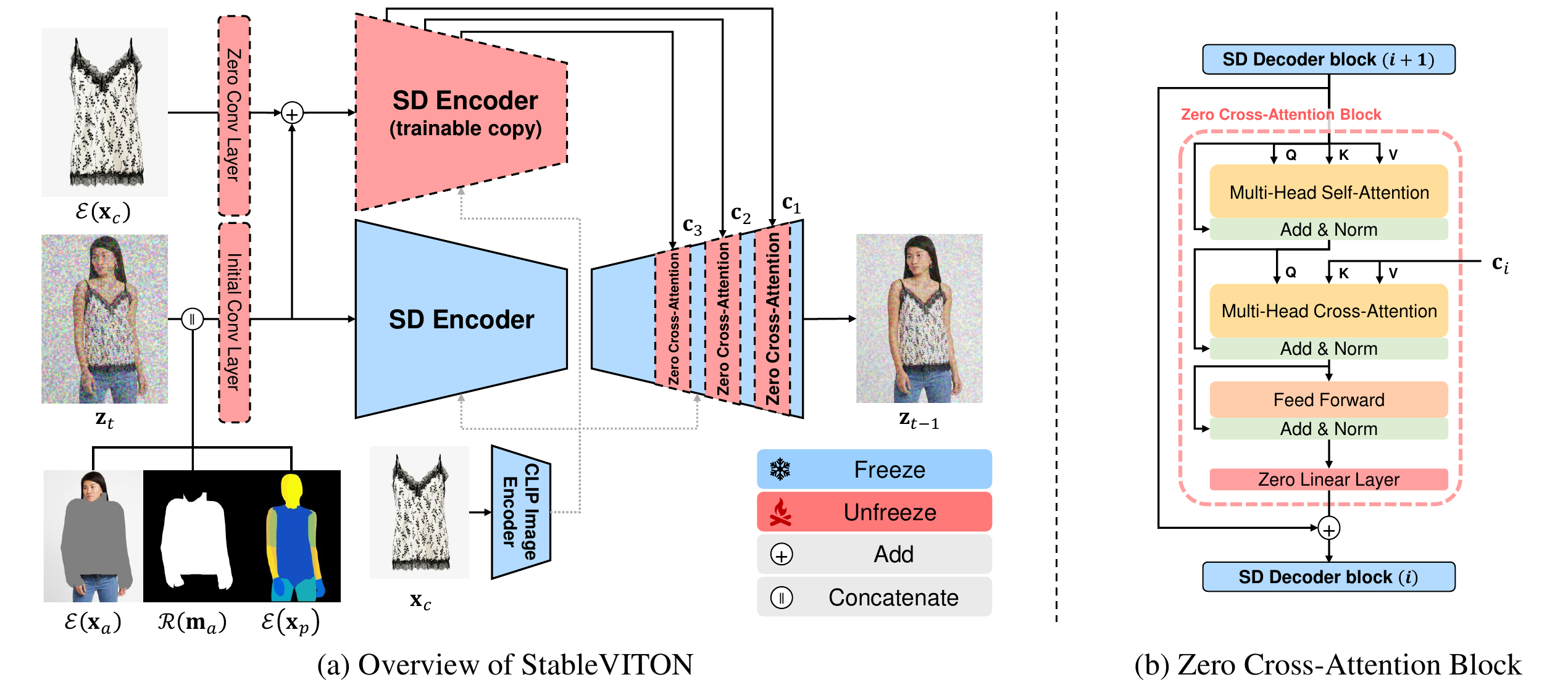}
    \vspace{-0.6cm}
    \caption{For the virtual try-on task,~\sysname~additionally takes three conditions: agnostic map, agnostic mask, and dense pose, as the input of the pre-trained U-Net, which serves as the query (Q) for the cross-attention. The feature map of the clothing is used as the key (K) and value (V) for the cross-attention and is conditioned on the UNet, as depicted in (b).}
    \vspace{-0.3cm}
    \label{fig:method_overview}
\end{figure*}

\section{Preliminary}
\noindent\textbf{Stable Diffusion Model.}
Stable Diffusion model~\cite{rombach2022high} is a large-scale diffusion model trained on LAION dataset~\cite{schuhmann2022laion}, built upon the Latent Diffusion model (LDM)~\cite{rombach2022high}, which performs a denoising process in the latent space of an autoencoder.
With a fixed encoder ($\mathcal{E}$), an input image $\mathbf{x}$ is first transformed to latent feature $\mathbf{z}_0 = \mathcal{E}(\mathbf{x})$. 
Given a pre-defined variance schedule $\beta_t$, we can define a forward diffusion process in the latent space following denoising diffusion probabilistic models~\cite{ho2020denoising}:
\begin{equation}
    q(\mathbf{z}_t | \mathbf{z}_0) = \mathcal{N}(\mathbf{z}_t ; \sqrt{\bar{\alpha}_t}\mathbf{z}_0, (1-\bar{\alpha}_t)\mathbf{I}),
\end{equation}
where $t \in \{1, ..., T\}$, $T$ represents the number of steps in the forward diffusion process, $\alpha_t:= 1-\beta_t$, and $\bar{\alpha}_t:= \Pi_{s=1}^t \alpha_s$. 
As a training loss, Stable Diffusion model employs the simplified objective function from LDM~\cite{lee2022high}: 
\begin{equation}
    \mathcal{L}_{LDM} = \mathbb{E}_{\mathcal{E}(\mathbf{x}),\mathbf{y},\epsilon\sim\mathcal{N}(0, 1),t}\left[\lVert\epsilon - \epsilon_{\theta}(\mathbf{z}_t, t, \tau_{\theta}(\mathbf{y}))\rVert_2^2\right],
\end{equation}
where the denoising network $\epsilon_{\theta}(\cdot)$ is implemented with a U-Net architecture~\cite{ronneberger2015u} and $\tau_{\theta}(\cdot)$ is the CLIP~\cite{radford2021learning} text encoder to condition the text prompt $y$.

\section{Method}
\subsection{Model Overview}
An overview of the~\sysname~is presented in Fig.~\ref{fig:method_overview}(a). 
Given a person image $\mathbf{x}\in\mathbb{R}^{H \times W \times 3}$, the clothing-agnostic person representation $\mathbf{x}_a \in \mathbb{R}^{H \times W \times 3}$ (we call it as `agnostic map')~\cite{choi2021viton} is proposed to eliminate any clothing information in $\mathbf{x}$. 
In this work, we approach the virtual try-on as an exemplar-based image inpainting problem~\cite{yang2023paint} to fill the agnostic map $\mathbf{x}_a$ with the clothing image $\mathbf{x}_c$. 
As the input of the U-Net, we concatenate four components: (1) the noisy image ($\mathbf{z}_t$), (2) latent agnostic map ($\mathcal{E}(\mathbf{x}_a)$), (3) the resized clothing-agnostic mask ($\mathbf{x}_{m_a}$), (4) latent dense pose condition ($\mathcal{E}(\mathbf{x}_p)$)~\cite{guler2018densepose} to preserve the person's pose.
To align the input channels, we expand the initial convolution layer of the U-Net to 13 (\textit{i.e.,} 4+4+1+4=13) channels with a convolution layer initialized with zero weights.
For exemplar conditioning, we input the $\mathbf{x}_c$ to the CLIP image encoder~\cite{yang2023paint}.

To preserve the fine details of the clothing, we introduce a spatial encoder, which takes latent clothing ($\mathcal{E}(\mathbf{x}_c)$) as input.
This spatial encoder copies the weight of the pre-trained U-Net~\cite{zhang2023adding} and conditions the intermediate feature maps of the encoder to U-Net via zero cross-attention blocks. 
During training, we apply augmentation and further finetune the model with our proposed attention total variation loss, which makes the attention region on the clothing sharper. The detailed model architecture is described in the supplementary material.

\subsection{\sysname}
\noindent\textbf{Zero Cross-Attention Block.}~\label{sec:zero_ca}
We aim to condition the intermediate feature maps of the clothing to U-Net, properly aligning with the human body.
The operation of adding the unaligned clothing feature map to the human feature map is insufficient to preserve clothing details due to misalignment between the human body and the clothing.
Therefore, we proposed a zero cross-attention block to be a flexible operation by applying an attention mechanism for conditioning.
Specifically, as shown in Fig.~\ref{fig:method_overview}(b), the feature map of the U-Net decoder block inputs to self-attention, followed by the cross-attention layer where the query (Q) comes from the previous self-attention layer and the spatial encoder's feature map serves as the key (K) and value (V). 
To eliminate harmful noise, we introduce a linear layer initialized with zero weight after the feed-forward operation~\cite{zhang2023adding}.

\begin{figure*}[t!]
    \centering
    \includegraphics[width=1\linewidth]{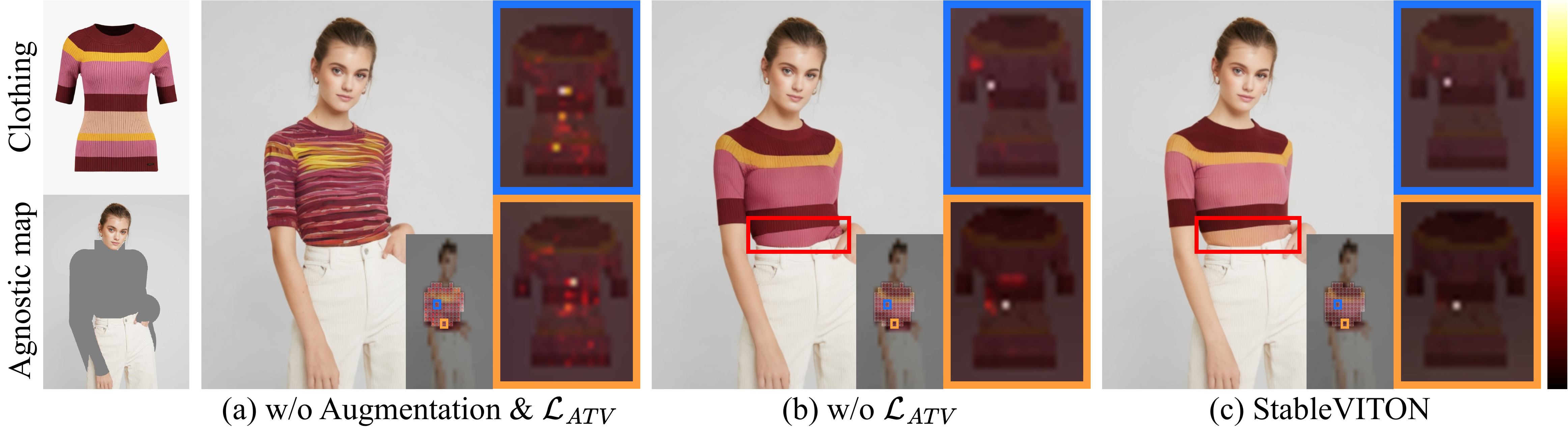}
    \vspace{-0.6cm}
    \caption{Visualization of attention map from a zero cross-attention block of $32\time24$ resolution.}
    \vspace{-0.3cm}
    \label{fig:attention_map}
\end{figure*}

To successfully align the clothing to the human body part via cross-attention, it is crucial to ensure semantic correspondence between the key tokens (clothing) and the query tokens (human body). For instance, when dealing with a query token related to the right shoulder, the corresponding key tokens should exhibit higher attention scores in the corresponding right shoulder area of the clothing. 
In Fig.~\ref{fig:attention_map}(a), we averaged the attention maps of the resolution of $32\times24$ across the head dimension and arranged them flatly. 
For clear visualization, we downsample the generated image to a resolution of $32\times24$ and then resize it back to $32^2\times24^2$. 
Subsequently, we overlay this generated image with the attention map corresponding to each query token.
Zooming in on the upper and middle of the generated clothing area, we observe that the key tokens unrelated to the corresponding query token, such as the bottom of the clothing, are activated in the attention map. This indicates that the cross-attention layer fails to learn the exact semantic correspondence between query and key tokens, combining the several key tokens of the clothing to generate the color corresponding to the query token during training. 
Therefore, as shown in Fig.~\ref{fig:attention_map}, the stripes on the clothing are not distinctly visible.

\noindent\textbf{Augmentation.} 
To mitigate such issues of key tokens unrelated to query tokens being attended to, we alter the feature map by applying augmentation, including random shifts to input conditions. 
Detailed settings of augmentation are described in supplementary material. 
Along with the augmented input conditions, we train our model with the objective function defined as follows:
\begin{equation}\label{eq:ldm_loss}
    \mathcal{L}_{LDM} = \mathbb{E}_{\mathbf{\zeta},\mathbf{x}_c, \mathcal{E}(\mathbf{x}_c),\epsilon,t}\left[\lVert\epsilon - \epsilon_{\theta}(\mathbf{\zeta}, t, \tau_{\phi}(\mathbf{x}_c), \mathcal{E}(\mathbf{x}_c))\rVert_2^2\right],
\end{equation}
where $\zeta=[\mathbf{z}_t;\mathcal{E}(\mathbf{x}_a);\mathbf{x}_{m_a};\mathcal{E}(\mathbf{x}_p)]$, and $\tau_{\phi}$ is the CLIP image encoder. Note that we do not update the parameters of the original blocks, as depicted in Fig.~\ref{fig:method_overview}(a).

The rationale behind this strategy is to force the model to learn fine-grained semantic correspondence using augmentation, instead of just moderately injecting the clothes at similar positions.
As shown in Fig.~\ref{fig:attention_map}(b), we can confirm that key tokens related to query tokens have high attention scores, signifying that the cross-attention layer has learned the high semantic correspondence between the clothing-agnostic region and clothing.

\noindent\textbf{Attention Total Variation Loss.}
While the cross-attention layer successfully aligns the clothing to the agnostic map, the points with high attention scores appear in dispersed positions, as shown in the attention map of Fig.~\ref{fig:attention_map}(b). 
This causes inaccurate details in generated images, such as color discrepancies. 

To address such an issue, we propose attention total variation loss. 
As the attention scores are the weight for the output, we calculate the center coordinates as a weighted sum of the attention map and the grid. 
Therefore, given the $H_q W_q$ query tokens and $h_k w_k$ key tokens, we calculate center coordinate map $F\in\mathbb{R}^{H_q \times W_q \times 2}$ as follows:
\vspace{-0.2cm}
\begin{equation}
    F_{ijn} = \frac{1}{h_k w_k}\sum_{k=1}^{h_k }\sum_{l=1}^{w_k }\left(A_{ijkl} \odot G_{kln}\right),
\end{equation}
where we average the attention map over the head dimension and reshape it as $A\in\mathbb{R}^{H_q \times W_q \times h_k \times w_k }$, and $G\in[-1,1]^{h_k \times w_k \times 2}$ is a 2D normalized coordinate. $\odot$ indicates element-wise multiplication operation.

For each query token in each clothing-agnostic region, the center coordinates should be evenly distributed, and the attention total variation loss $\mathcal{L}_{ATV}$ is defined as follows:
\begin{equation}
    \mathcal{L}_{ATV} = \parallel\nabla(F \odot M)\parallel_1,
\end{equation}
where $M\in\{0,1\}^{H_q \times W_q}$ is the ground truth clothing mask to only affect the clothing region. 
The attention total variation loss  $\mathcal{L}_{ATV}$ is designed to enforce the center coordinates on the attention map uniformly distributed, thereby alleviating interference among attention scores located at dispersed positions. As illustrated in Figure (c), this leads to the generation of a sharper attention map, thereby more accurately reflecting the color of the clothing. 

Finally, we finetune our~\sysname~ by adding $\mathcal{L}_{ATV}$ to Eq.~\ref{eq:ldm_loss}:
\vspace{-0.1cm}
\begin{equation}
    \mathcal{L}_{finetune} = \mathcal{L}_{LDM} + \lambda_{ATV}\mathcal{L}_{ATV},
\end{equation}
where $\lambda_{ATV}$ is a weight hyper-parameter.

%% file: 4_experiment.tex
\section{Experiment}
\noindent\textbf{Baselines.}
We compare~\sysname~with three GAN-based virtual try-on methods, VITON-HD~\cite{choi2021viton}, HR-VITON~\cite{lee2022high}, and GP-VTON~\cite{xie2023gp}, and two diffusion-based virtual try-on methods, LADI-VTON~\cite{morelli2023ladi} and DCI-VTON~\cite{gou2023taming}. 
We also evaluate a diffusion-based inpainting method, Paint-by-Example~\cite{yang2023paint}. 
We use pre-trained weights if available; otherwise, we train the models following the official code. 

\begin{figure*}[t!]
    \centering
    \includegraphics[width=1\linewidth]{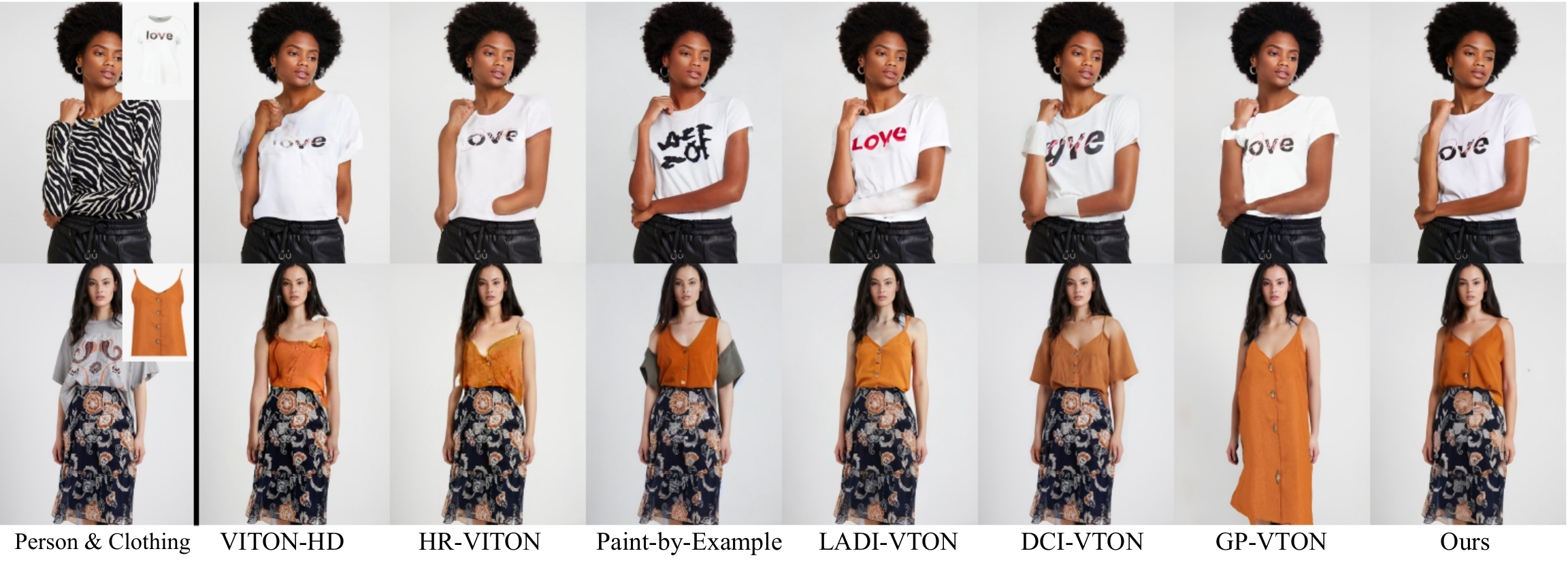}
    \vspace{-0.8cm}
    \caption{Qualitative comparison with baselines in a single dataset setting (VITON-HD / VITON-HD). Best viewed when zoomed in.}
    \label{fig:qual_indomain}
    \vspace{-0.2cm}
\end{figure*}

\begin{figure*}[t!]
    \centering
    \includegraphics[width=1\linewidth]{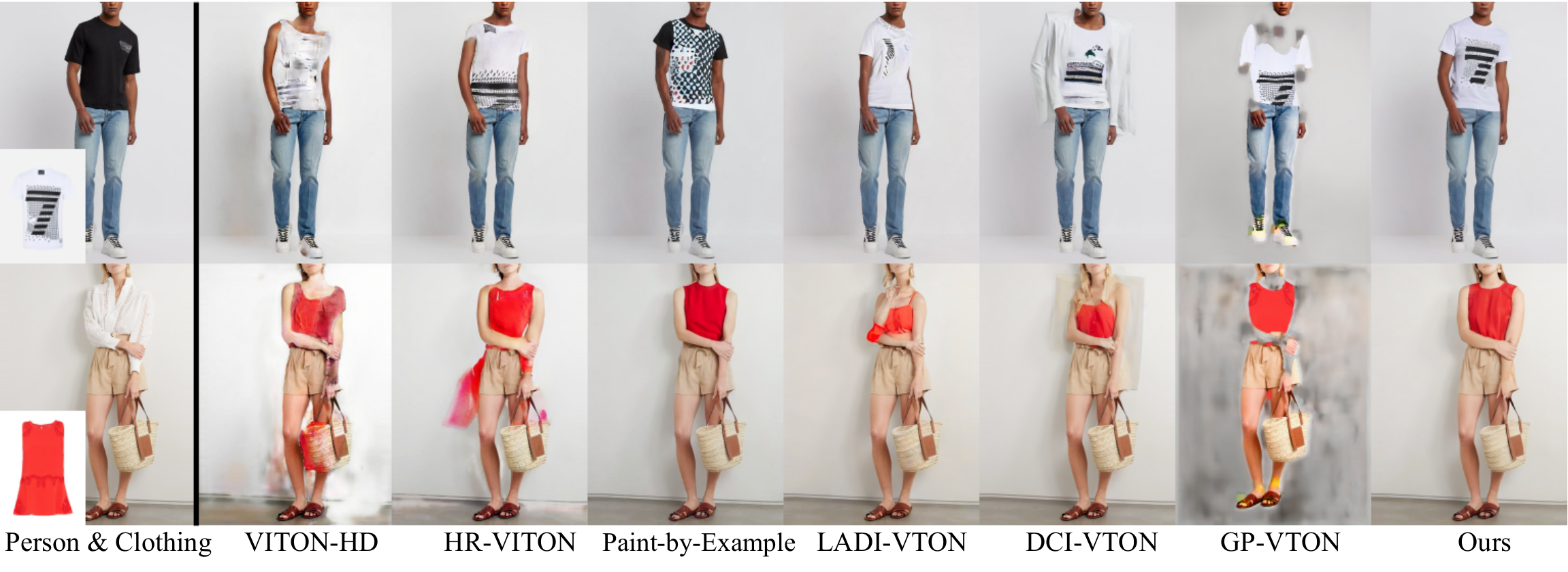}
    \vspace{-0.8cm}
    \caption{Qualitative comparison with baselines in a cross dataset setting (VITON-HD / DressCode). Best viewed when zoomed in.}
    \label{fig:qual_crossdomain1}
    \vspace{-0.2cm}
\end{figure*}

\begin{table*}[t!]
\centering
\resizebox{0.60\linewidth}{!}{
\begin{tabular}{lcccclcccc} 
\toprule
\textbf{Train / Test}   &
\multicolumn{4}{c}{\textbf{VITON-HD / VITON-HD}} &  & \multicolumn{4}{c}{\textbf{D.C. Upper / D.C. Upper}}              \\ 
\cline{2-5}\cline{7-10}
\textbf{Method~} &
\textbf{SSIM}  & \textbf{LPIPS}  & \textbf{FID}   & \textbf{KID}  &  &
\textbf{SSIM}  & \textbf{LPIPS}  & \textbf{FID}  & \textbf{KID}   \\ 
\hline
\textbf{VITON-HD~\cite{choi2021viton}} &
0.862          & 0.117           & 12.117         & 3.23          &  &
-              & -               & -              & -              \\
\textbf{HR-VITON~\cite{lee2022high}} &
0.878          & 0.1045          & 11.265         & 2.73          &  &
\underline{0.936}          & 0.0652          & 13.820         & 2.71           \\
\textbf{LADI-VTON~\cite{morelli2023ladi}} &
0.864          & 0.0964          & 9.480          & 1.99          &  &
0.915          & 0.0634          & 14.262         & 3.33           \\
\textbf{Paint-by-Example~\cite{yang2023paint}} &
0.802          & 0.1428          & 11.939         & 3.85          &  &
0.897          & 0.0775          & 15.332         & 4.64           \\
\textbf{DCI-VTON~\cite{gou2023taming}} &
0.880          & \underline{0.0804}          & 8.754          & 1.10          &  &
\textbf{0.937} & \underline{0.0421}          & 11.920         & 1.89           \\
\textbf{GP-VTON~\cite{xie2023gp}} &
\underline{0.884}          & 0.0814          & 9.072          & \underline{0.88}          &  &
0.769          & 0.2679          & 20.110         & 8.17           \\ 
\hline
\textbf{Ours} &
0.852          & 0.0842          & \underline{8.698}          & \underline{0.88}          &  &
0.911          & 0.0500          & \underline{11.266}         & \underline{0.72}           \\
\textbf{Ours~(RePaint~\cite{lugmayr2022repaint})} &
\textbf{0.888} & \textbf{0.0732} & \textbf{8.233} & \textbf{0.49} &  &
\textbf{0.937} & \textbf{0.0388} & \textbf{9.940} & \textbf{0.12}  \\
\bottomrule
\end{tabular}}
\vspace{-0.3cm}
\caption{Quantitative comparisons in single dataset settings, VITON-HD and DressCode upper-body (D.C. Upper) datasets. \textbf{Bold} and \underline{underline} denote the best and the second best result, respectively.}
\vspace{-0.35cm}
\label{tab:quan_in_domain}
\end{table*}

\noindent\textbf{Dataset.}
We conduct the experiments using two publicly available virtual try-on datasets, VITON-HD~\cite{choi2021viton} and DressCode~\cite{morelli2022dress}, and one human image dataset, SHHQ-1.0~\cite{fu2022styleganhuman}. 

We train our model with VITON-HD and upper-body images in DressCode, respectively.
For the evaluation of SHHQ-1.0, we use the first 2,032 images and follow the preprocessing instruction of VITON-HD~\cite{choi2021viton} to obtain the input conditions such as the agnostic maps or the dense pose. 

\noindent\textbf{Evaluation.}
We evaluate the performances in two test settings.
Specifically, the paired setting uses a pair of a person and the original clothes for reconstruction, whereas the unpaired setting involves changing the clothing of a person image with a different clothing item.
As previous work~\cite{choi2021viton}, training and evaluation within a single dataset are referred to as `single dataset evaluation'.
On the other hand, we extend our evaluation on other datasets (\textit{e.g.}, SHHQ-1.0), which we refer to as a `cross dataset evaluation'.
This evaluation enables an in-depth assessment of the model's generalizability in handling arbitrary person images, demonstrating applicability in real-world scenarios.
Our model is capable of training at a $1024\times768$ resolution, but for a fair evaluation with baselines, we used a model trained at a $512\times384$ resolution.
More results and details about experiments are described in the supplementary material.

\subsection{Qualitative Results}
\noindent\textbf{Single Dataset Evaluation.}
As shown in Fig.~\ref{fig:qual_indomain},~\sysname~generates realistic images and effectively preserves the text and clothing textures compared to the six baseline methods.
Specifically, in the first row of Fig.~\ref{fig:qual_indomain}, GAN-based methods such as GP-VTON struggle to generate the arms of the target person naturally.
Moreover, other diffusion-based models either fail to preserve the text (Paint-by-Example and LADI-VTON) or show an overlapped artifact between the clothing and the target person (DCI-VTON). 
On the other hand, despite some parts of the arm being covered by clothing, our model produces a high-fidelity result that omits the `L' in `Love'.

\begin{figure*}[t!]
    \centering
    \includegraphics[width=1\linewidth]{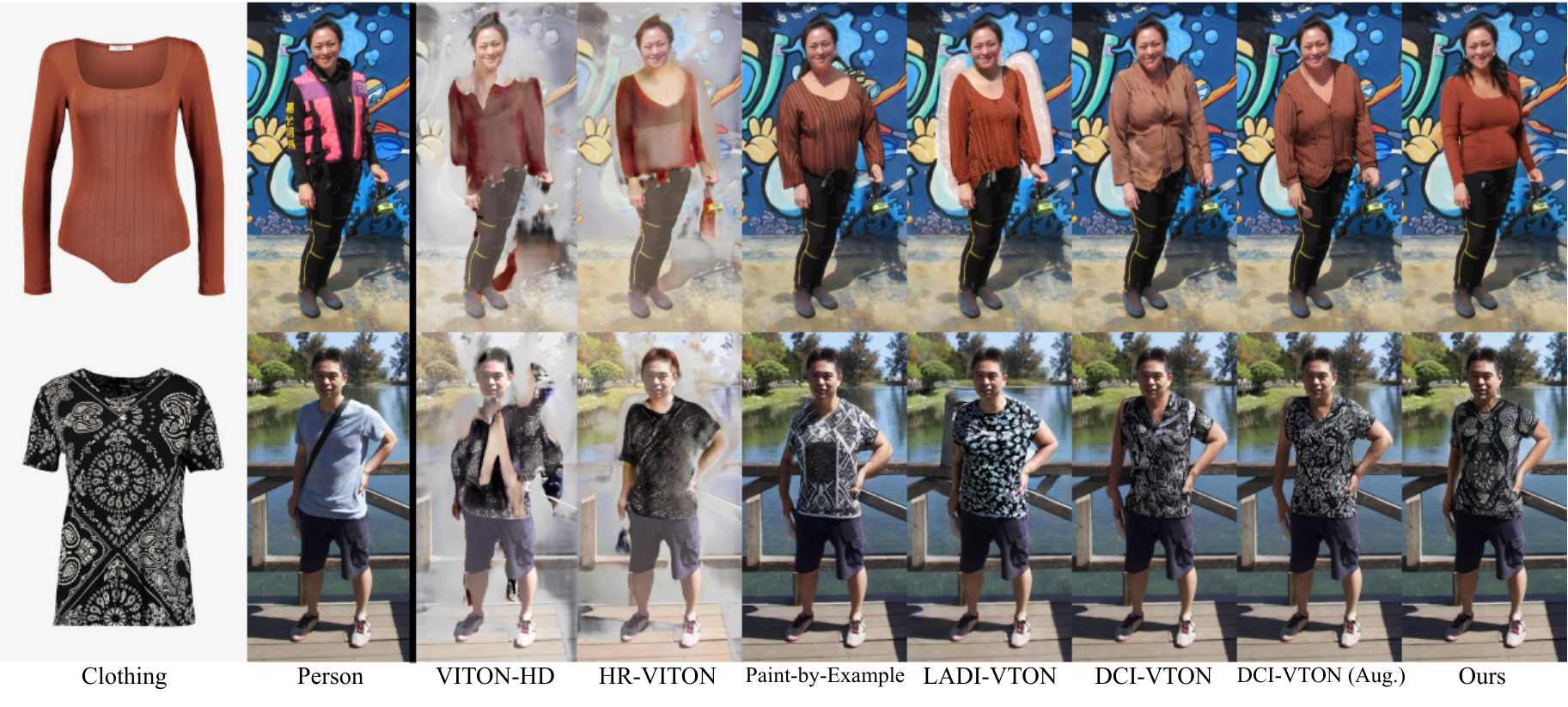}
    \vspace{-0.7cm}
    \caption{Qualitative comparison with baselines in a cross dataset setting (VITON-HD / SHHQ-1.0). Best viewed when zoomed in.}
    \vspace{-0.2cm}
    \label{fig:qual_crossdomain2}
\end{figure*}

\noindent\textbf{Cross Dataset Evaluation.}
We visualize the generation images of the models trained on VITON-HD for DressCode and SHHQ-1.0 datasets in Fig.~\ref{fig:qual_crossdomain1} and Fig.~\ref{fig:qual_crossdomain2}, respectively.
The results clearly demonstrate that~\sysname~generates high-fidelity images while preserving the details of the clothing. 
GAN-based methods especially show significant artifacts on the target person and fail to maintain background. 
While diffusion-based methods generate natural images, they fail to preserve clothing details or the shape of the clothing. 
Furthermore, even when applying the augmentation we used to DCI-VTON (denoted as DCI-VTON (Aug.)), as depicted in Fig.~\ref{fig:qual_crossdomain2}, a significant improvement in the performance of the warping network is not achieved, failing to preserve clothing details.

\begin{table*}[t!]
\centering
\resizebox{0.8\linewidth}{!}{
\begin{tabular}{lcccccccccclcc}
\toprule
\textbf{Train / Test}  &  &
\multicolumn{4}{c}{\textbf{VITON-HD / D.C. Upper}} &  & \multicolumn{4}{c}{\textbf{D.C. Upper / VITON-HD}} &  & \multicolumn{2}{c}{\textbf{VITON-HD / SHHQ-1.0}}  \\ 
\cline{3-6}\cline{8-11}\cline{13-14}
\textbf{Method} &  &
\textbf{SSIM}  & \textbf{LPIPS}  & \textbf{FID}    & \textbf{KID}  &  &
\textbf{SSIM}  & \textbf{LPIPS}  & \textbf{FID}   & \textbf{KID}  &  &
\textbf{FID}   & \textbf{KID}                    \\ 
\hline
\textbf{VITON-HD~\cite{choi2021viton}} &  & 
0.853  & 0.1874 & 44.257 & 28.82  &  &
-      & -      & -      & -      &  & 
71.149 & 52.01 \\
\textbf{HR-VITON~\cite{lee2022high}} &  &
0.909  & 0.1077 & 19.970 & 7.35 &  &
0.811  & 0.2278 & 45.923 & 36.69 &  &
52.732 & 31.22  \\
\textbf{LADI-VTON~\cite{morelli2023ladi}} &  &
0.901  & 0.1009 & 16.336 & 5.36 &  &
0.801  & 0.2429 & 31.790 & 23.02 &  &
24.904 & 6.07 \\
\textbf{Paint-by-Example~\cite{yang2023paint}} &  &
0.889   & 0.0867 & 16.398 & 4.78 &  &
0.784   & 0.1814 & 15.625 & 7.52 &  & 
26.274  & 9.830 \\
\textbf{DCI-VTON~\cite{gou2023taming}} &  &
0.903             & 0.1217 & 23.076 & 12.03 &  &
\underline{0.825} & 0.1870 & 16.670 & 6.40 &  &
24.850 & 6.68 \\
\textbf{DCI-VTON (Aug.)~\cite{gou2023taming}} &  &
0.898  & 0.1240 & 18.809 & 8.02 &  &
-      & -      & -      & -    &  &
24.368 & 6.11 \\
\textbf{GP-VTON~\cite{xie2023gp}} &  &
0.724 & 0.3846 & 65.711 & 66.01 &  &
0.804 & 0.2621 & 52.351 & 48.68 &  &
-     & - \\ 
\hline
\textbf{Ours} &  &
\underline{0.911}  & \underline{0.0603} & \underline{12.581} & \underline{1.70} &  &
0.817              & \underline{0.1308} & \underline{10.104} & \underline{1.72} &  &
\underline{23.531} & \underline{5.68} \\
\textbf{Ours~(RePaint~\cite{lugmayr2022repaint})} &  & 
\textbf{0.938}  & \textbf{0.0470} & \textbf{10.480} & \textbf{0.41} &  &
\textbf{0.855}  & \textbf{0.1173} & \textbf{9.714}  & \textbf{1.35} &  &
\textbf{21.077} & \textbf{5.10} \\
\bottomrule
\end{tabular}}
\vspace{-0.3cm}
\caption{Quantitative comparisons in cross dataset settings. We train the models on VITON-HD and DressCode upper-body (D.C. Upper) datasets and evaluate them on different datasets. \textbf{Bold} and \underline{underline} denote the best and the second best result, respectively.}
\vspace{-0.35cm}
\label{tab:quan_cross_domain}
\end{table*}

\subsection{Quantitative Results}
\noindent\textbf{Metrics.}
For quantitative evaluation, we use SSIM~\cite{wang2004image} and LPIPS~\cite{zhang2018unreasonable} in the paired setting. 
In an unpaired setting, we assess the realism using FID~\cite{heusel2017gans} and KID~\cite{binkowski2018demystifying} score. 
We follow the evaluation paradigm~\cite{morelli2023ladi} for the implementation~\cite{detlefsen2022torchmetrics,parmar2022aliased}.

\noindent\textbf{Single Dataset Evaluation.}
We evaluate our~\sysname~and existing baselines on a single dataset setting and report the results in Table~\ref{tab:quan_in_domain}. 
In the unpaired setting (\textit{i.e.,} FID and KID),~\sysname~outperforms all the baselines. 
We observe that the performance degradation in the paired setting occurs due to the autoencoder's reconstruction error of the agnostic map. 
To mitigate this issue, we adapt RePaint~\cite{lugmayr2022repaint}, which samples the known region (\textit{i.e.}, agnostic map) and replaces it in each denoising steps during the inference, used in DCI-VTON. 
Applying RePaint,~\sysname~outperforms the baselines for all evaluation metrics. 
Since RePaint greatly helps maintain regions unrelated to clothing, it shows notable performance improvement in the paired setting.
Nevertheless, even without RePaint, our method demonstrates superior performance in terms of FID and KID in the unpaired setting compared to the baselines.

\noindent\textbf{Cross Dataset Evaluation.}
Table~\ref{tab:quan_cross_domain} presents that our~\sysname~shows state-of-the-art performance for all the evaluation metrics with a large margin. 
GAN-based methods fail to maintain background consistency, resulting in significantly high FID and KID scores.
While diffusion-based methods exhibit better performance due to the plausible generation outcomes enabled by pre-trained diffusion models, they fail to preserve clothing details. Consequently, they exhibit lower similarity scores in the paired setting (\textit{i.e.,} SSIM and LPIPS) compared to our method.

\begin{figure}[t!]
    \centering
    \includegraphics[width=1\linewidth]{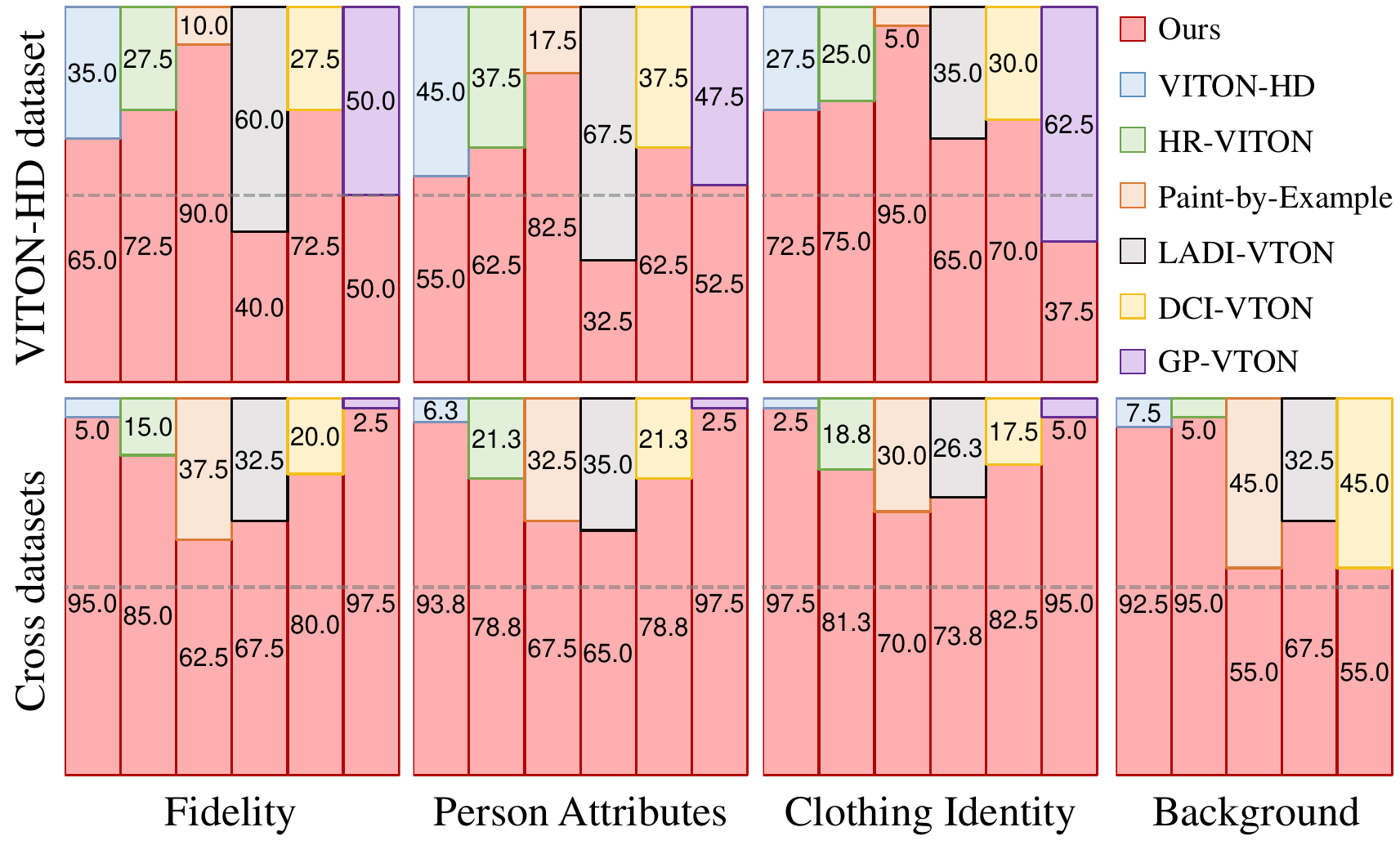}
    \vspace{-0.65cm}
    \caption{User study results. We compare our~\sysname~with six baselines, involving a total of 40 participants.}
    \vspace{-0.40cm}
    \label{fig:qual_userstudy}
\end{figure}

\subsection{User Study}
For the models trained on the VITON-HD dataset, we conducted a user study with 40 participants. 
Each participant was shown one image generated by the baseline and the other by our model. 
They were asked to choose the better image based on three criteria: (1) fidelity, (2) person attributes, and (3) clothing identity. 
We added a question about (4) background quality for the cross dataset setting. 
Detailed questions can be found in the supplementary material. 
As shown in Fig.~\ref{fig:qual_userstudy},~\sysname~outperforms in most of the criteria, and especially in the cross dataset setting, our method is overwhelming in human evaluations.
While LADI-VTON shows better preference in the evaluation of fidelity and person attributes on the VITON-HD dataset, it fails to preserve clothing details, resulting in a 35\% preference in the clothing identity criteria. 

%% file: 5_ablation.tex
\begin{table}[h!]
\centering
\resizebox{0.85\linewidth}{!}{
\begin{tabular}{l|cccc} 
\toprule
\textbf{Train / Test on VITON-HD}       & \textbf{SSIM} & \textbf{LPIPS} & \textbf{FID} & \textbf{KID}  \\ 
\hline
\textbf{ControlNet + Aug.}           & 0.832         & 0.1157         & 9.81         & 1.81          \\
\textbf{ControlNet-W + Aug.}         & 0.822         & 0.1124         & 9.66         & 1.57          \\
\textbf{Zero Cross-Attention Block + Aug.} & \textbf{0.850}         & \textbf{0.0851}         & \textbf{8.74}         & \textbf{0.91}          \\
\bottomrule
\end{tabular}}
\vspace{-0.2cm}
\caption{Quantitative comparison results between our zero cross-attention block and ControlNet. We train ControlNet with clothing and warped clothing~\cite{lee2022high} (ControlNet-W). We apply augmentation to all the models in training.}
\vspace{-0.5cm}
\label{tab:abl_controlnet}
\end{table}

\subsection{Ablation Study}
\begin{figure}[t!]
    \centering
    \includegraphics[width=1\linewidth]{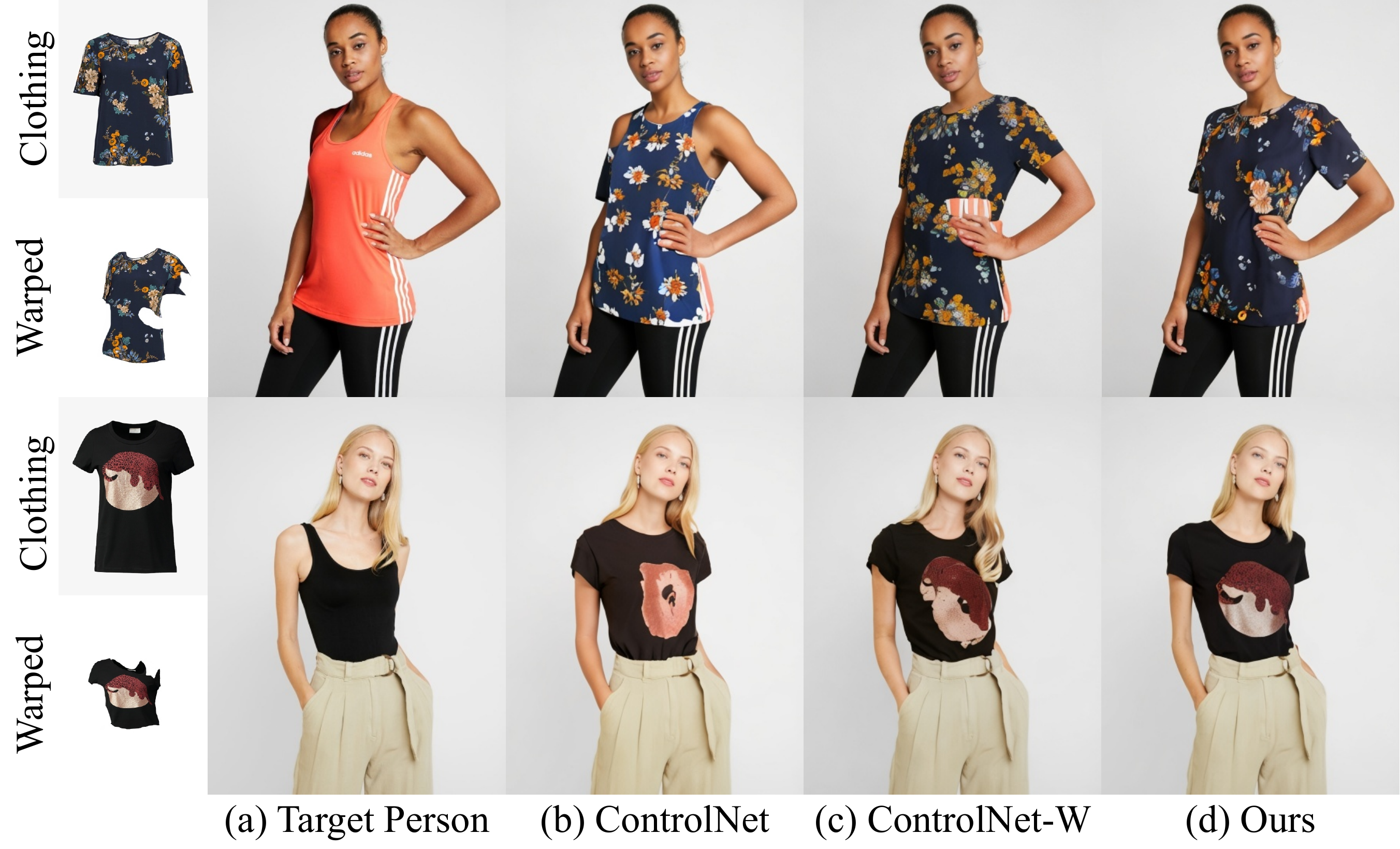}
    \vspace{-0.7cm}
    \caption{Comparison between our~\sysname~ and ControlNets under the two different input conditions: (1) clothing, and (2) warped clothing~\cite{lee2022high} (ControlNet-W). Best viewed when zoomed in.}
    \vspace{-0.3cm}
    \label{fig:abl_controlnet}
\end{figure}

\vspace{-0.2cm}
\noindent\textbf{Comparison with ControlNet.}
To demonstrate the effectiveness of~\sysname~in tackling the alignment issue compared to ControlNet~\cite{zhang2023adding}, we train ControlNet under two different input conditions: (1) clothing, and (2) warped clothing~\cite{lee2022high} (dubbed as ControlNet-W). 
We apply our proposed augmentation to both models during training. 
Using the warping network to align the clothing helps ControlNet capture coarse features, such as the overall shape and color of the logo, as shown in Fig.~\ref{fig:abl_controlnet}(b) and (c). 
However, since the misalignment still exists between warped clothing and the human body in the training phase, ControlNet-W struggles to reflect more fine-grained clothing details to the generation results. 
These observations highlight that ControlNet is highly sensitive to subtle misalignment across the input, stemming from the limitations of the ControlNet's direct addition operation in conditioning. 
In contrast, as shown in Fig.~\ref{fig:abl_controlnet}(d), the zero cross-attention block, free from the alignment constraints, successfully preserves the logos and patterns and leads to a qualitative performance improvement, as shown in Table~\ref{tab:abl_controlnet}. 

\begin{figure}[t!]
    \centering
    \includegraphics[width=1\linewidth]{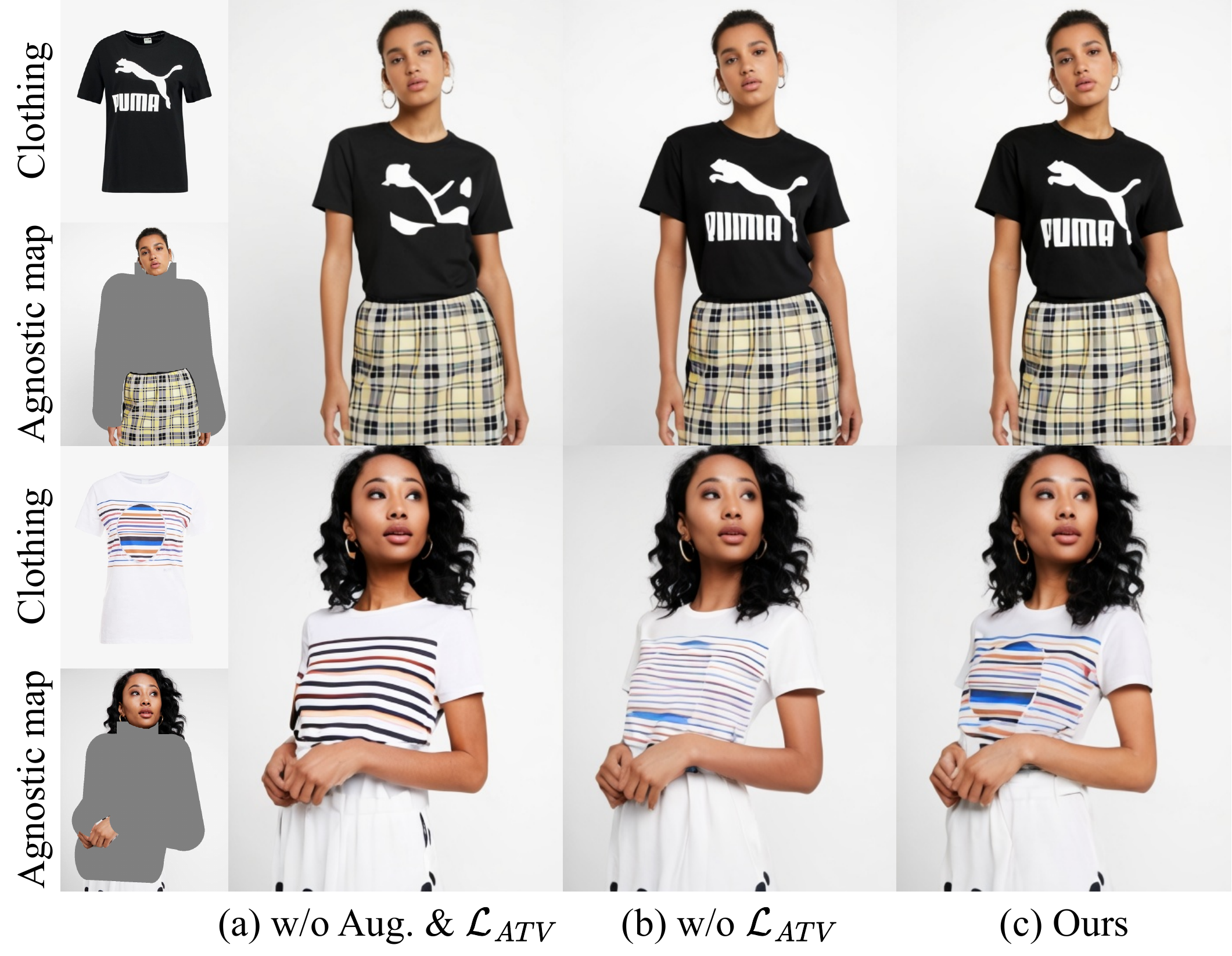}
    \vspace{-0.7cm}
    \caption{Visual comparisons according to our training components. Best viewed when zoomed in.}
    \vspace{-0.3cm}
    \label{fig:abl_component}
\end{figure}

\noindent\textbf{Effect of Training Components.}
We investigate the effect of the two proposed components during training: augmentation and attention total variation loss. 
In Fig.~\ref{fig:abl_component}, we visualize the generated images while incrementally introducing the proposed training components one by one. 
Compared to Fig~\ref{fig:abl_component}(a) and (b), we observe that detailed features such as logos and patterns of the clothing are more preserved when augmentation is applied. 
However, as the attention maps are not sufficiently distinct, we observe inaccuracies in the generated images, such as the `M' in `PUMA' being incorrectly depicted or lines blurring, as shown in Fig.~\ref{fig:abl_component}(b). 
After finetuning with our proposed attention total variation loss, these finer details are significantly improved.
Such visual enhancements correspond to quantitative performance improvements as demonstrated in Table~\ref{tab:abl_component}.

\begin{table}
\centering
\resizebox{0.70\linewidth}{!}{
\begin{tabular}{cc|cccc} 
\toprule
\textbf{Aug.} & \begin{tabular}[c]{@{}l@{}}\textbf{Attention}\\\textbf{TV Loss}\end{tabular} & \textbf{SSIM} & \textbf{LPIPS} & \textbf{FID} & \textbf{FID}  \\ 
\hline
\ding{55}              & \ding{55}                                                                            & 0.847         & 0.0969         & 9.35         & 1.33          \\
\ding{51}             & \ding{55}                                                                           & 0.850         & 0.0851         & 8.744         & 0.91          \\
\ding{51}             & \ding{51}                                                                            & \textbf{0.852}         & \textbf{0.042}          & \textbf{8.698}        & \textbf{0.88}          \\
\bottomrule
\end{tabular}}
\vspace{-0.25cm}
\caption{Ablation study of our proposed training components on VITON-HD dataset.}
\vspace{-0.35cm}
\label{tab:abl_component}
\end{table}

%% file: 6_conclusion.tex
\section{Conclusion}
We propose~\sysname, a novel image-based virtual try-on method using the pre-trained diffusion model. 
Our proposed zero cross-attention block learns semantic correspondence between the clothing and the human body, enabling try-on in the latent feature space. 
A novel attention total variation loss and augmentation are designed to preserve the clothing details better. 
Extensive experiments, including cross dataset evaluation, clearly demonstrate that~\sysname~shows the state-of-the-art performance compared to the existing methods and its promising quality in the real-world setting.

\noindent\textbf{Acknowledgements} Sunghyun Park is the corresponding author.

%% file: 7_suppl.tex
\maketitlesupplementary
\appendix
\setcounter{figure}{10}

\section{Implementation details}
\noindent\textbf{Architecture Details.}
We adopt the autoencoder and the denoising U-Net of the Stable Diffusion v1.4~\cite{rombach2022high}.
We initialize our denoising U-Net with the weights of the U-Net from the Paint-by-Example~\cite{yang2023paint}. 
The U-Net's encoder and decoder both consist of 12 blocks, involving three downsampling and upsampling steps each. 
As a result, when~\sysname~receives the input image of $64\times48$ resolution (after going through the autoencoder), three intermediate feature maps are generated for each of the following resolutions: $8\times6$, $16\times12$, $32\times24$, $64\times48$.
We use the feature maps at resolutions other than $8\times6$ as inputs for each of the nine zero cross-attention blocks. 
Similarly, for the spatial encoder following the U-Net's encoder structure, we utilized the nine feature maps at resolutions other than $8\times6$ as the key and value inputs for the cross-attention layers. 

\noindent\textbf{Training \& Inference Details.}
We train the model using an AdamW optimizer with a fixed learning rate of 1e-4 for 360k iterations, employing a batch size of 32. 
Then, we finetune the model with the attention total variation weight hyper-parameter $\lambda_{ATV}=0.001$, using the same learning rate and batch size for 36K iterations. 
We train for about 100 hours using four NVIDIA A100 GPUs.
For augmentation, we simultaneously applied horizontal flip (p=0.5) to both the clothing and the UNet's input condition, and independently applied Random Shift (limit=0.2, p=0.5) and Random Scale (limit=0.2, p=0.5) to both the clothing and UNet's input. 
We simultaneously applied HSV adjustments (limit=5, p=0.5) and contrast adjustments (limit=0.3, p=0.5) to both the clothing and $\mathbf{x}_0$. 
To prevent the issue of facial distortion, we finetune the decoder of the autoencoder separately on the training datasets VITON-HD~\cite{choi2021viton} and DressCode~\cite{morelli2022dress}. 
In training the decoder, we use the AdamW optimizer with a learning rate of 5e-5 and a batch size of 32 for 10k iterations on each dataset. 
For inference, we employ the pseudo linear multi-step (PLMS)~\cite{liu2022pseudo} sampler, with the number of sampling steps set to 50.

\section{User Study Details.}
In the user study, participants were asked to evaluate which of the two images, one generated by the baseline and the other by StableVITON, was superior in terms of 1) fidelity, 2) person attributes, 3) clothing identity, and 4) background quality (with respect to the cross-dataset setting). The questions for each criterion were as follows:

\begin{itemize}
    \item Fidelity: Choose which image better exhibits resemblance to reality in aspects such as the human body and color harmony. 
    \item Person Attributes: Choose which image better maintains features like skin tone, pose, and appearance from the input image.
    \item Clothing Identity: Choose which image better preserves characteristics such as the design, logo, and shape of the input clothing.
    \item Background Quality: Choose which image better maintains the background of the input image.
\end{itemize}

\section{Additional Qualitative Results}
\begin{figure*}[t!]
    \centering
    \includegraphics[width=1\linewidth]{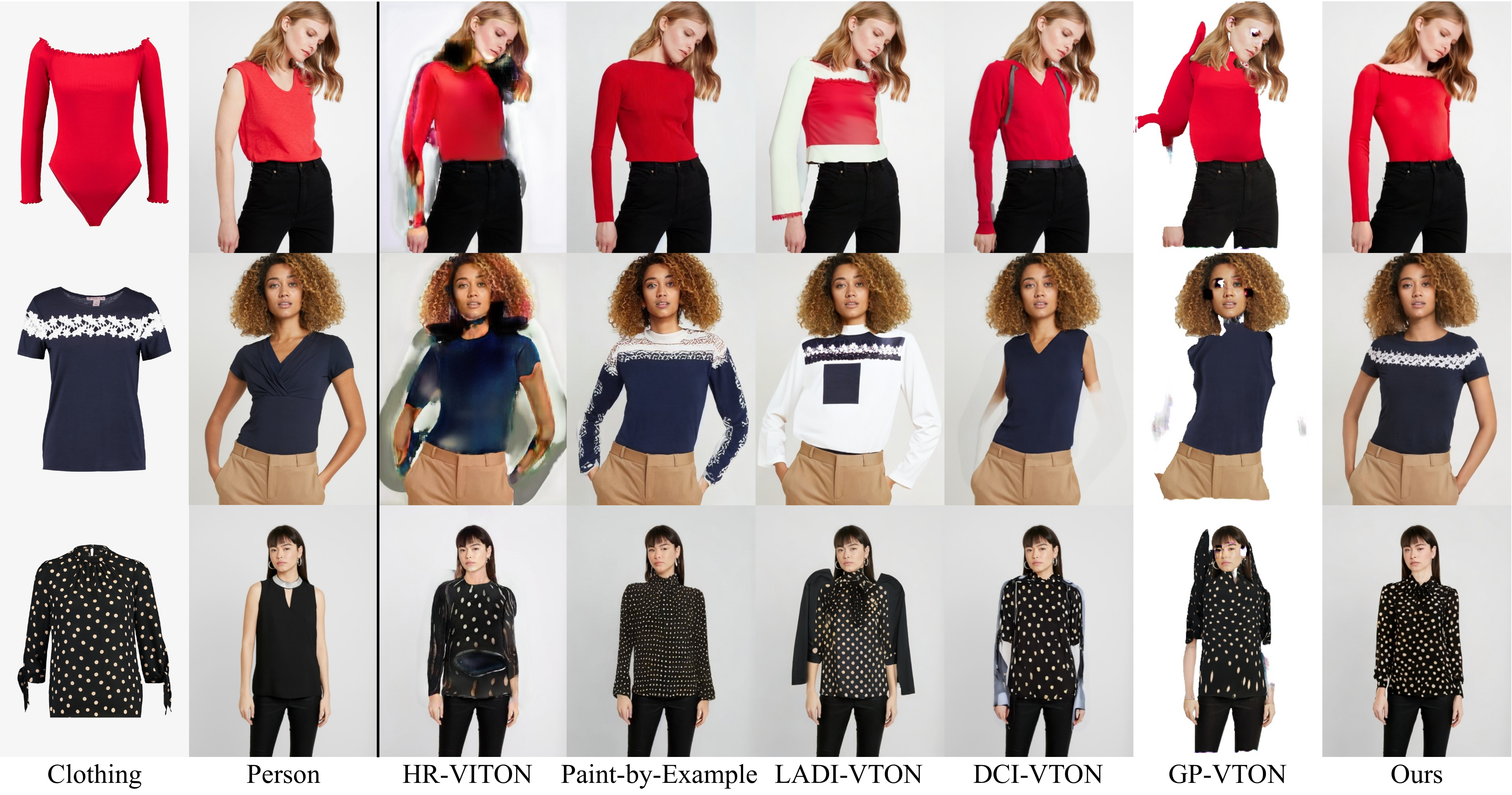}
    \caption{Qualitative comparison with baselines in a cross dataset setting (DressCode / VITON-HD). Best viewed when zoomed in.}
    \label{fig:supple_cross_domain}
\end{figure*}

In Fig.~\ref{fig:supple_cross_domain}, we present the generation results on VITON-HD dataset, using the models trained on DressCode upper body dataset. GAN-based models (\textit{i.e.}, HR-VITON and GP-VTON) show significant artifacts around the target person, as evidenced in quantitative results (see Table 2 in the main paper), leading to high FID and KID scores. In addition, diffusion-based models, while providing a plausible appearance, fail to preserve the details of the clothing. 

\section{\sysname~at High Resolution}
To synthesize high-fidelity images, we have further trained~\sysname~at the higher resolution of $1024\times768$.
Instead of starting from scratch, we experimentally observed that progressively training~\sysname~with $1024\times768$ resolution images leads to faster convergence. 
We conducted additional training for 85k iterations using the same training settings as~\sysname.

\noindent\textbf{Qualitative Results.}
We present the results generated by~\sysname~trained on VITON-HD dataset at $1024\times768$ resolution for images in VITON-HD, DressCode, SHHQ-1.0, and web-crawled datasets, in Fig.~\ref{fig:supple_viton_grid}, Fig.~\ref{fig:supple_dresscode_grid}, Fig.~\ref{fig:supple_shhq_grid}, and Fig.~\ref{fig:supple_web_grid}, respectively. 
We observe that there is a clearer preservation of facial or clothing details at $1024\times768$ resolution.

\section{Limitations \& Discussion}
Even when fine-tuning the decoder of the autoencoder on the virtual try-on dataset, it remains challenging to preserve very fine details of the face or clothing. 
As a result, as shown in Fig.~5 of the main paper, subtle variations in facial features, such as the eyes, can be observed. 
However, we effectively address these issues related to fine details by increasing the model's resolution, as demonstrated in Fig.~\ref{fig:supple_viton_grid}.

In our experiments, we observed that our model fails to preserve objects occluding the person or accessories such as bracelets and watches attached to the target person. 
This issue arises from the model's inability to incorporate additional information, apart from clothing, during the sampling process to fill the masked regions of the agnostic map. 
We leave such preservation issues as future work.

\clearpage

\begin{figure*}
    \centering
    \includegraphics[width=1\linewidth]{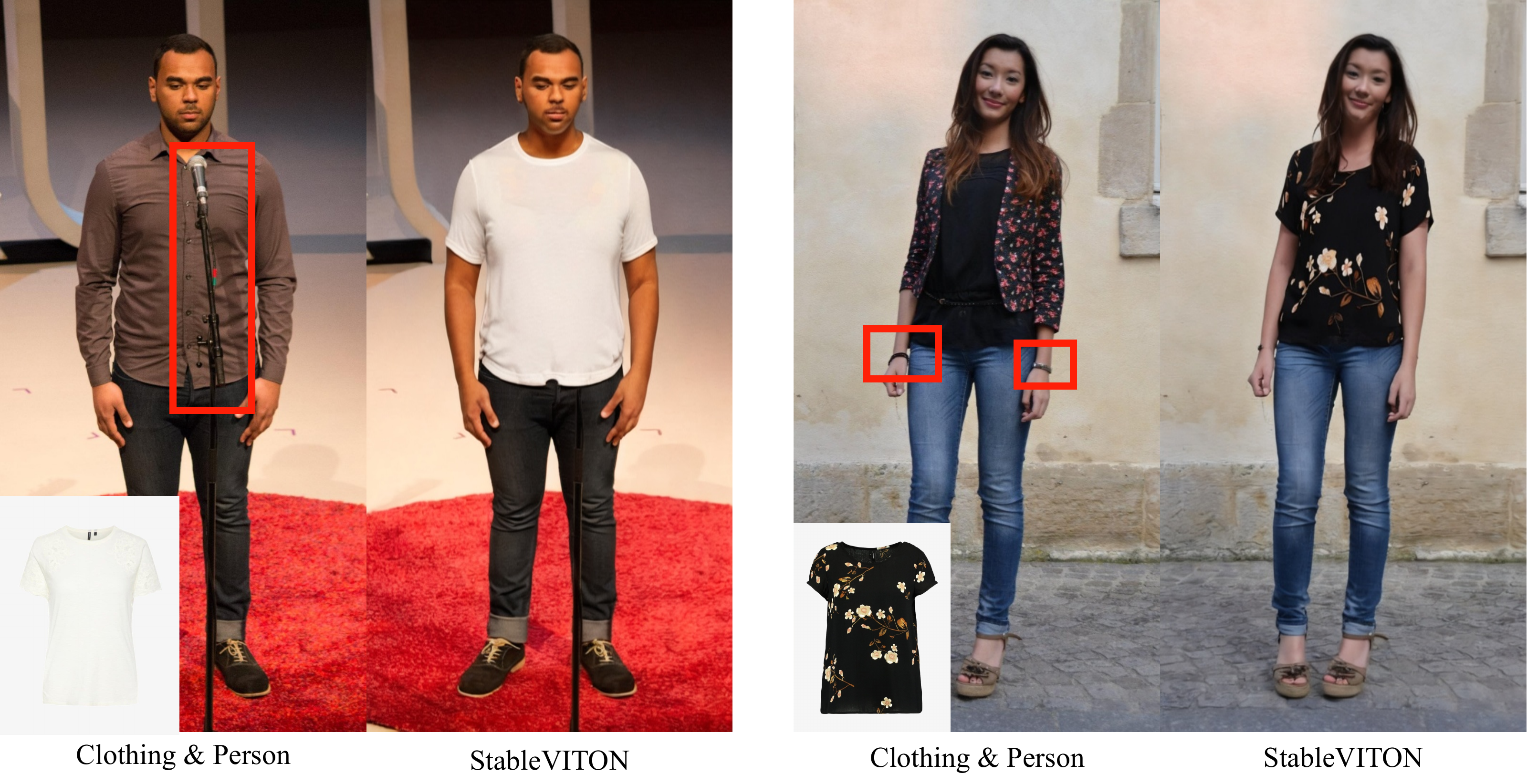}
    \caption{Limitations of~\sysname.~\sysname~fails to preserve objects occluding the person or accessories such as bracelets.}
    \label{fig:supple_limitation}
\end{figure*}

\begin{figure*}[t!]
    \centering
    \includegraphics[width=1\linewidth]{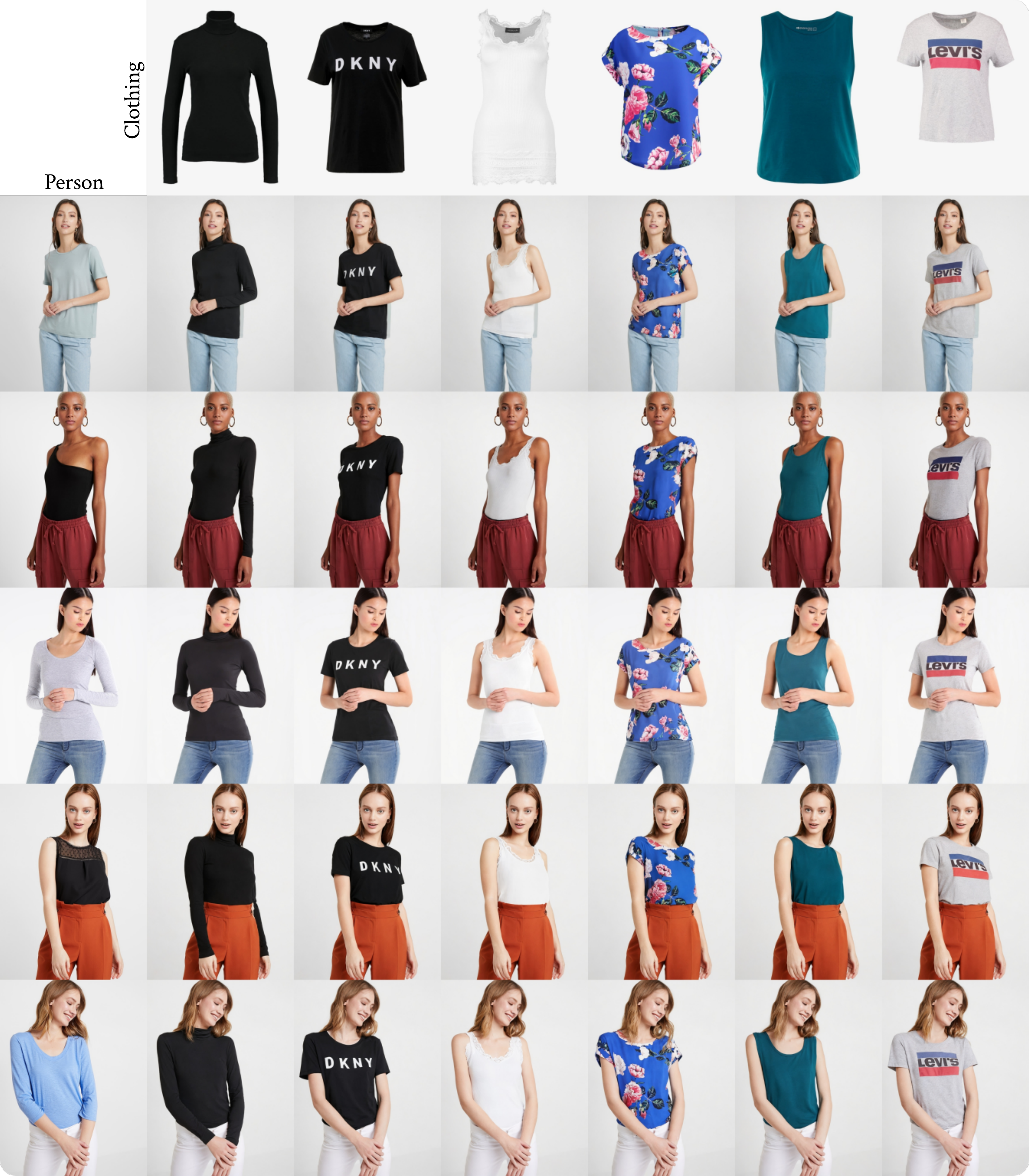}
    \caption{The generation results for the VITON-HD test dataset by~\sysname~trained on VITON-HD dataset at $1024\times768$ resolution. Best viewed when zoomed in.}
    \label{fig:supple_viton_grid}
\end{figure*}

\begin{figure*}[t!]
    \centering
    \includegraphics[width=1\linewidth]{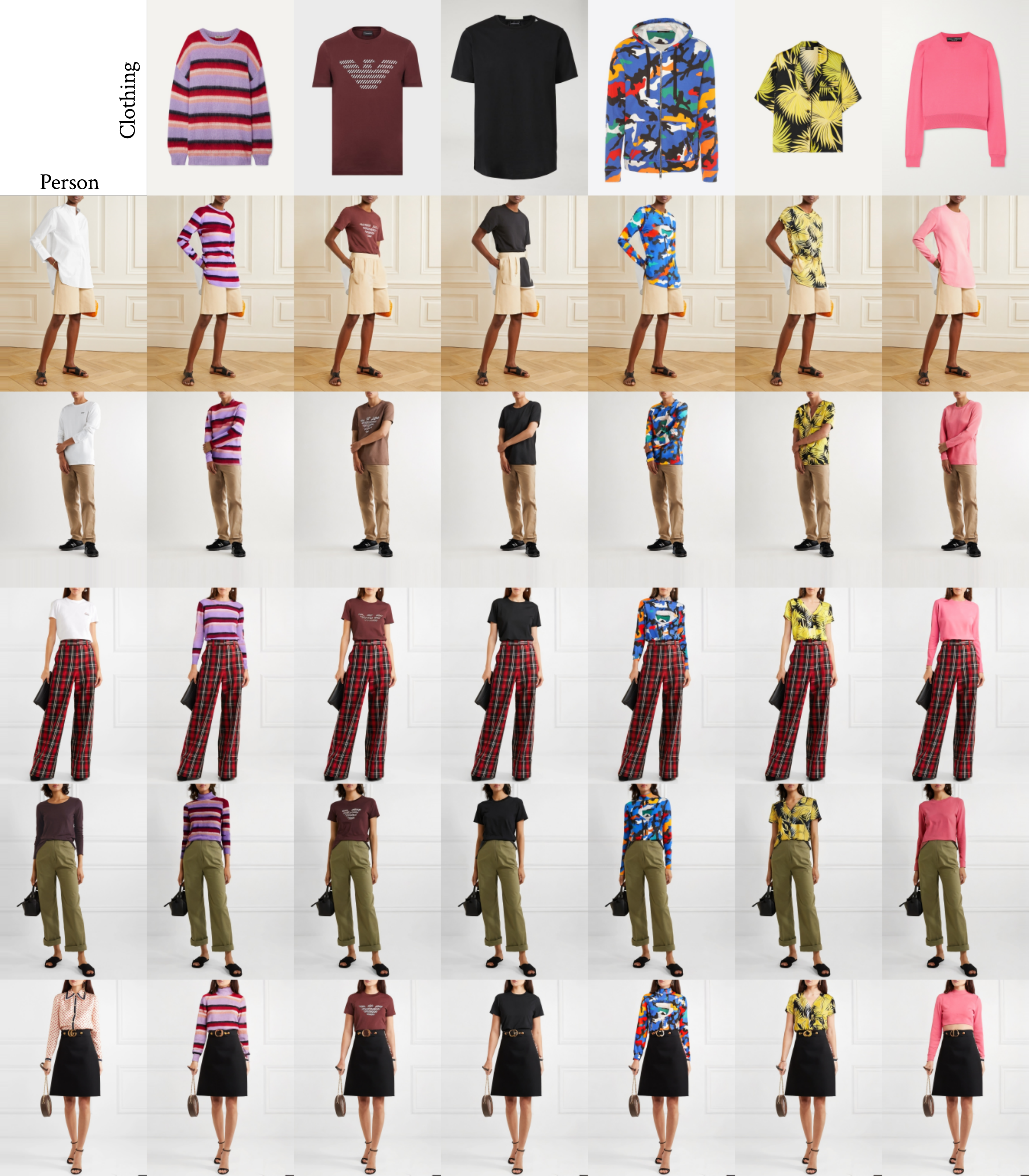}
    \caption{The generation results for the DressCode dataset by~\sysname~trained on VITON-HD dataset at $1024\times768$ resolution. Best viewed when zoomed in.}
    \label{fig:supple_dresscode_grid}
\end{figure*}

\begin{figure*}[t!]
    \centering
    \includegraphics[width=1\linewidth]{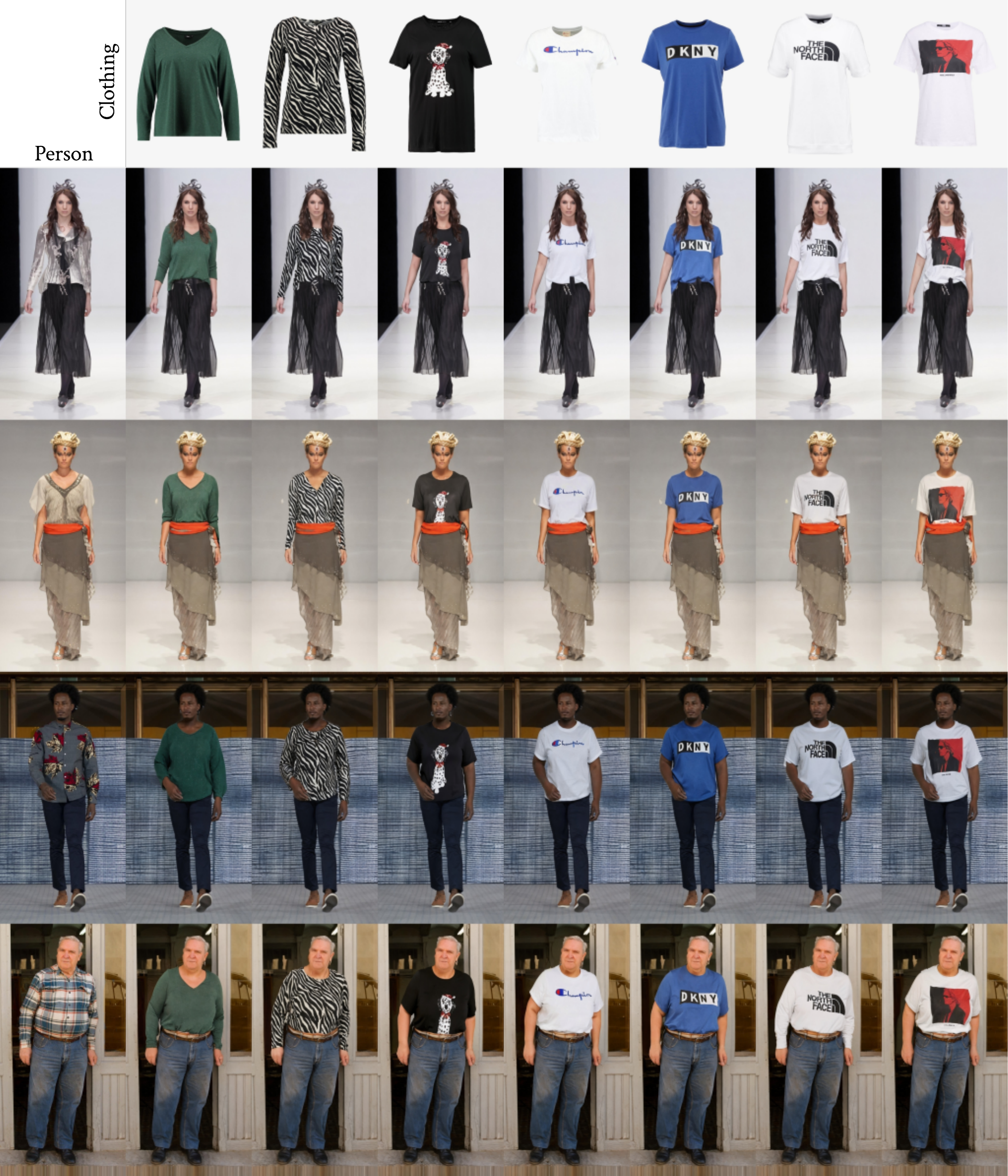}
    \caption{The generation results for the SHHQ-1.0 dataset by~\sysname~trained on VITON-HD dataset at $1024\times768$ resolution. Best viewed when zoomed in.}
    \label{fig:supple_shhq_grid}
\end{figure*}

\begin{figure*}[t!]
    \centering
    \includegraphics[width=1\linewidth]{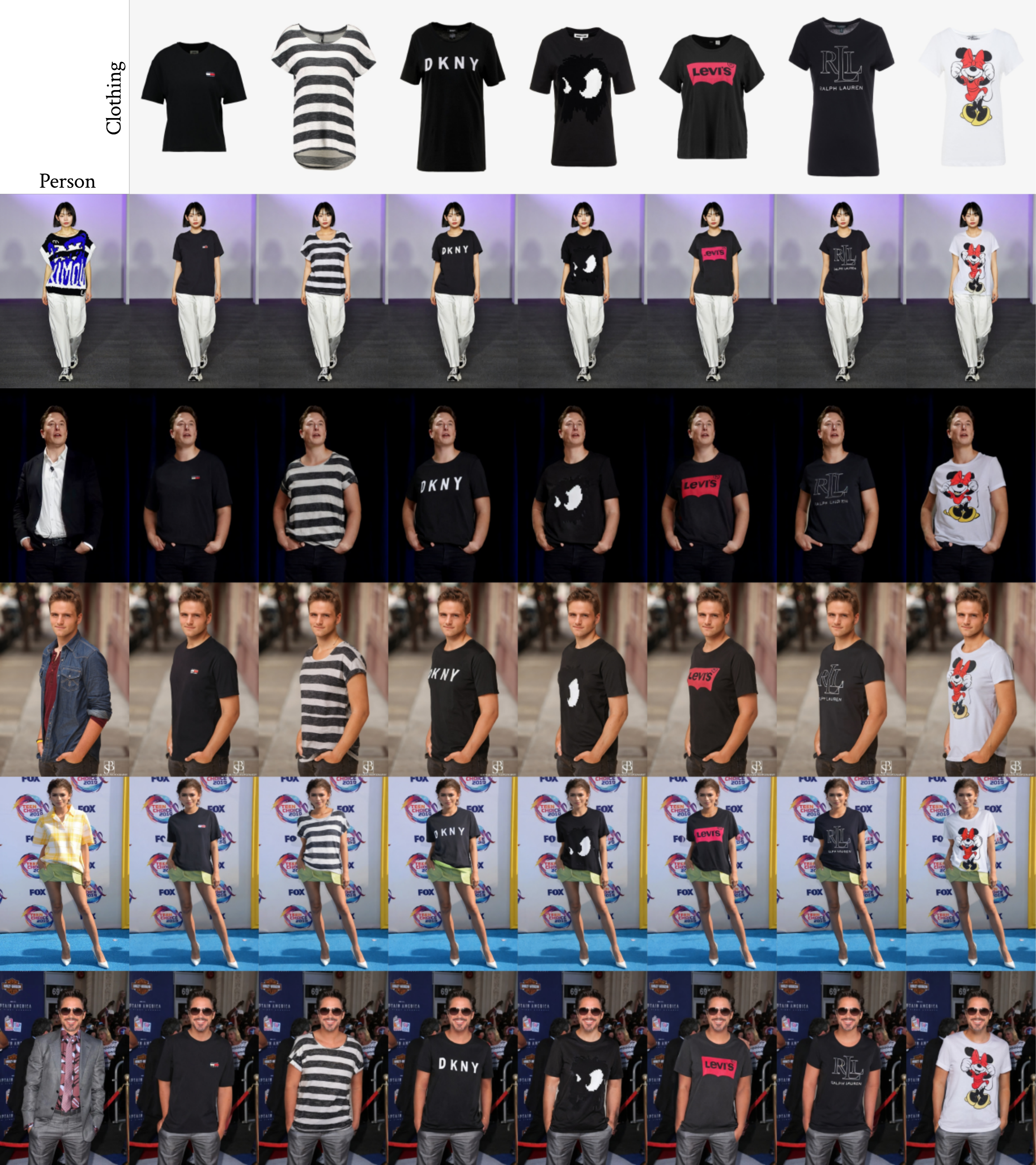}
    \caption{The generation results for the web-crawled images by~\sysname~trained on VITON-HD dataset at $1024\times768$ resolution. Best viewed when zoomed in.}
    \label{fig:supple_web_grid}
\end{figure*}
\clearpage